\documentclass{article}
\usepackage{ijcai26}

\usepackage{times}
\usepackage{soul}
\usepackage{url}
\usepackage[hidelinks]{hyperref}
\usepackage[utf8]{inputenc}
\usepackage[small]{caption}
\usepackage{graphicx}
\usepackage{amsmath}
\usepackage{amsthm}
\usepackage{booktabs}
\usepackage{algorithm}
\usepackage{algorithmic}
\usepackage[switch]{lineno}
\usepackage{newfloat}
\usepackage{listings}
\usepackage{amssymb}

\title{Bridging the phenotype-target gap for molecular generation via multi-objective reinforcement learning}

\author{
Haotian Guo
\and
Hui Liu$^{*}$\\
\affiliations
College of Computer and Information Engineering, Nanjing Tech University, Nanjing, 211800, China.\\
\emails
hliu@njtech.edu.cn
}

\begin{document}

\maketitle

\begin{abstract}
   The generation of high-quality candidate molecules remains a central challenge in AI-driven drug design. Current phenotype-based and target-based strategies each suffer limitations, either incurring high experimental costs or overlook system‑level cellular responses. To bridge this gap, we propose XMolRL, a novel generative framework that synergistically integrates phenotypic and target-specific cues for de novo molecular generation. The phenotype-guided generator is first pretrained on expansive drug‑induced transcriptional profiles and subsequently fine‑tuned via multi‑objective reinforcement learning (RL). Crucially, the reward function fuses docking affinity and drug‑likeness scores, augmented with ranking loss, prior-likelihood regularization, and entropy maximization. The multi‑objective RL steers the model toward chemotypes that are simultaneously potent, diverse, and aligned with the specified phenotypic effects. Extensive experiments demonstrate XMolRL's superior performance over state-of-the-art phenotype-based and target-based models across multiple well-characterized targets. Our generated molecules exhibit favorable drug-like properties, high target affinity, and inhibitory potency (IC50) against cancer cells. This unified framework showcases the synergistic potential of combining phenotype-guided and target-aware strategies, offering a more effective solution for de novo drug discovery. The source code and datasets are available at: \url{https://github.com/hliulab/XMolRL}.
\end{abstract}

\section{Introduction}
In AI-driven drug design, generating high-quality candidate molecules remains a fundamental challenge~\cite{sanchez2018inverse,pang2023deep}. In recent years, deep generative models have made significant advances in molecular design~\cite{gomez2018automatic,aini2024antimalarial,brown2019guacamol}. Most existing methods largely fall into two distinct paradigms: target-based and phenotype-based drug discovery. Target-based approaches mainly focus on well-characterized targets—such as receptors, enzymes and transport proteins~\cite{hughes2011principles,imming2006drugs,wishart2018drugbank}—and aim to design molecules that specifically bind to active sites or binding pockets. These methods leverage structural information to guide rational drug design and elucidate mechanisms of action. Nevertheless, their effectiveness relies heavily on the availability of accurate target structures and well-understood biological functions~\cite{swinney2011were,polykovskiy2020molecular,von2020exploring} . In addition, target-based methods often overlook system-level cellular responses, limiting their ability to capture off-target effects and broader phenotypic consequences.

In contrast, phenotype-based strategies seek to identify bioactive compounds that elicit desired phenotypic changes in cellular systems~\cite{vincent2022phenotypic,moffat2014phenotypic}. This paradigm does not require prior knowledge of biological targets, allowing for a more holistic view of drug effects in complex cellular contexts. This is especially valuable when the underlying molecular mechanisms of a disease are poorly understood~\cite{moffat2017opportunities,meissner2022emerging}. However, large-scale experiments to observe drug-induced, system-level phenotypic effects remain prohibitively expensive and labor-intensive.~\cite{swinney2011were}. Furthermore, these experiments are difficult to standardize or automate, limiting the throughput of phenotype-based drug screening~\cite{moffat2017opportunities}. Fortunately, transcriptional profiles effectively capture how compounds perturb the cellular state, offering molecular-level readouts that reflect drug perturbations leading to higher-order phenotypes~\cite{lamb2006connectivity}. Due to their high throughput and cost-efficiency, gene expression signatures have been widely used to probe both genetic alterations and external stimuli that lie along the causal pathway between genotype and disease phenotype. In fact, differential expression signatures have been employed to identify compounds capable of inducing desired phenotypic shifts~\cite{Subramanian2017L1000}. Accordingly, we use the terms ``phenotypic profile'' standing for the transcriptional profiles in this study.

To date, few efforts have leveraged the complementary strengths of target-based and phenotype-based paradigms. To bridge this gap, we propose XMolRL, a generative framework that integrates phenotypic and target-specific information for de novo molecular generation. XMolRL comprises two key components: 1) a phenotype-guided molecular generator trained on large-scale drug-induced phenotypic profiles enabling the conditional generation of molecules that elicit desired drug efficacy; and 2) a reinforcement learning module that fine-tunes the generator using docking-based affinity scores to guide optimization toward specific protein targets. To ensure robust and diverse molecular generation, we incorporate multiple designed regularization terms into the RL objective, to mitigate reward exploitation and maintain molecular diversity and drug-likeness. Our extensive experiments demonstrate that XMolRL can generate candidate molecules that simultaneously exhibit desired phenotypic effects, strong target affinity, and favorable drug-like properties. This unified framework showcases the synergistic potential of combining phenotype-driven and structure-aware strategies, offering a more comprehensive and effective solution for de novo drug discovery. Our main contributions are summarized as below: 
\begin{itemize}
    \item We propose a novel framework that integrates phenotypic profiles with target protein structures to jointly optimize molecular efficacy and specificity.
    \item We introduce ranking loss, prior likelihood, and entropy regularization to enhance stability, diversity, and overall generation quality in reinforcement learning.
    \item Our method achieves superior performance over state-of-the-art phenotype-based and target-based models across multiple targets, simultaneously excelling in binding affinity, drug-likeness, and synthetic feasibility.
\end{itemize}

\section{Related Works}
\subsection{Phenotype-Guided Molecular Generation}
Phenotype-based molecular generation focuses on leveraging cellular-level response data—such as transcriptional profiles—to guide the design of compounds capable of modulating disease states~\cite{vincent2022phenotypic,moffat2014phenotypic}. This approach is particularly valuable for complex or poorly understood diseases where molecular mechanisms remain elusive. With the integration of deep learning and omics technologies, phenotype-driven drug discovery models have seen rapid development. For example, PaccMann~\cite{cadow2020paccmann} predicts drug sensitivity by modeling the relationship between gene expression profiles of cancer cell lines and compound SMILES representations, capturing phenotype-to-structure associations. TRIOMPHE~\cite{kaitoh2021triomphe} employs a variational autoencoder conditioned on expression profiles to generate candidate molecules. Gex2SGen~\cite{das2023gex2sgen} and GxVAEs~\cite{li2024gxvaes} followed similar ideas that used the desired expression profiles as input to design drug-like molecules capable of eliciting phenotypic changes. Mendez-Lucio et al.~\cite{mendez2020novo} further employed a generative adversarial network (GAN) in conjunction with expression profiles to automatically design molecules with a high probability to induce therapeutic effects. SmilesGEN~\cite{liu2025phenotypic} is a dual-channel variational autoencoder framework that jointly models drug molecules and gene expression profiles in a shared latent space to generate drug-like molecules capable of eliciting desired phenotypic effects.

\subsection{Target-based Molecular Design}
Target-driven approaches take advantage of protein structure or binding pocket information to generate molecules with high binding affinity~\cite{danel2023docking}. 
Representative autoregressive models, such as SBDD-3D~\cite{luo20213d} and Pocket2Mol~\cite{peng2022pocket2mol}, treat ligand generation as a stepwise 3D construction process. By encoding the binding pocket as geometric point clouds, these methods leverage rotationally invariant or equivariant networks to iteratively sample atoms and bonds, ensuring the generated structures strictly adhere to the specific binding site geometry. 
LiGAN~\cite{mendez2020novo} and AutoGrow4~\cite{spiegel2020autogrow4} leverage the three‑dimensional geometry of protein binding pockets together with fragment‑assembly heuristics to generate molecules that satisfy spatial docking constraints. By contrast, OptiMol~\cite{nigam2021beyond} and SampleDock~\cite{xu2021navigating} employ a ``generate-evaluate-select'' workflow: they sample candidates from a learned latent space, score each via molecular docking, retain the highest‑scoring compounds, and iteratively refine the generator based on this feedback. Moreover, SBMolGen~\cite{ma2021structure} integrates reinforcement learning by treating docking scores as rewards, thereby steering the generative policy toward progressively more favorable chemotypes. 
Nevertheless, drug efficacy is governed by the intricate cellular environment, and molecules optimized exclusively for target binding frequently fail to produce the intended phenotypic outcomes.

\begin{figure*}
    \centering
    \includegraphics[width=1\linewidth]{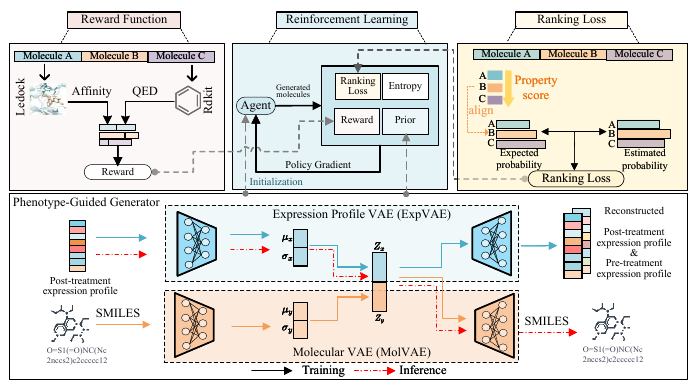}
    \caption{Overview of the XMolRL architecture. The model consists of a pretrained phenotypic-profile-guided generator, while a reinforcement learning agent fine-tunes the molecular generation toward desired biochemical properties.}
    \label{fig:enter-label}
\end{figure*}
\section{XMolRL Framework}
We present XMolRL, a generative framework that integrates multi-modal data—including phenotypic profiles and chemical language—for the generation of drug-like molecules. XMolRL comprises a phenotypic profile-guided generative module, and a reinforcement learning module that fine-tunes the agent toward desired molecular properties such as binding affinity and quantitative estimate of drug-likeness (QED) scores. The phenotypic generator is pretrained on large-scale drug-induced expression profiles and then acts as a prior model. Next, RL is utilized to refine the agent model toward desired molecular attributes, while mitigating reward hacking effects that may compromise molecular uniqueness. Once trained, XMolRL generates molecules guided by expression profiles and informed by the structural context of target proteins.

\subsection{Phenotype-Guided Generator}
The phenotype-guided generator aims to produce drug-like molecules conditioned on desired transcriptional profiles. Inspired by SmilesGEN~\cite{liu2025phenotypic}, we adopt a dual-channel variational autoencoder (VAE) architecture, which consists of ExpVAE, a multi-layer feedforward network used to encode expression profiles, and MolVAE, a GRU-backboned VAE designed to encode and reconstruct molecular SMILES sequences. MolVAE is firstly pre-trained on a large-scale SMILES dataset, after which its encoder is frozen. Next, joint training is performed using triplets comprising SMILES sequences, perturbed expression profiles, and corresponding unperturbed profiles. This allows for the simultaneous reconstruction of both molecular structures and expression profiles, leading to joint optimization of the parameters in ExpVAE and MolVAE.

The dual-VAE architecture learns a latent space that captures the complex relationship between phenotypic effects and chemical structures, enhancing its capacity to generate biologically relevant molecules. During inference, we retain the encoder from ExpVAE and the decoder from MolVAE to enable phenotypic drug design. The trained generator serves as a prior model in RL phase, constraining the policy to produce molecules aligned with desired phenotypic effect.

\subsection{RL-Based Joint Molecular Generation}
To jointly optimize phenotypic efficacy and target binding, we cast phenotype and target-based molecular generation as a reinforcement learning (RL) problem. Specifically, the trained phenotypic generator serves as a prior, and an agent model adopts the same network architecture. During the RL phase, the agent is initialized with the prior's parameters and optimized to maximize the task-specific reward function. Notably, the prior remains fixed throughout the RL phase and serves as a regularizer, constraining the agent’s policy to generate molecules that induce biologically plausible phenotypic effects and preserving consistency with phenotype-conditioned distribution learned during pretraining.

To guide the generation of molecules with desirable properties, we incorporate both molecule-target docking scores and QED values into the reward function to encourage the agent to explore promising regions of the chemical space. The binding affinity between each generated molecule and its target protein is quantified using LeDock docking scores, a tool that has demonstrated strong performance in a recent benchmark study involving 2002 protein–ligand complexes ~\cite{zhao2013discovery,wang2016comprehensive}. The QED score is computed using RDKit and incorporated into the reward function. Let  \( S = \{s_1, s_2, \dots, s_n\} \) denote a set of SMILES sequences, where each \( s \in S \) is of the form \( s = \{a_1, a_2, \dots, a_t\} \), with \( a \) denoting a token in the sequence.
For each SMILES sequence $s$, the reward function is defined as follows:
\begin{equation}
    Reward(s)=Dock(s)\times QED(s)
\end{equation}
where $Dock$ denotes the normalized docking score and QED is the drug-likeness score of the generated molecule. A molecule is considered \underline{valid} only if it satisfies RDKit's validity check and its QED score exceeds a predefined threshold. We set the QED threshold to 0.34, which represents the mean score of compounds classified as 'unattractive' or 'too complex'~\cite{bickerton2012quantifying}.  If a generated molecule is deemed invalid (e.g., chemically implausible or syntactically incorrect), its docking score is set to zero as a form of penalty. Moreover, we have observed that imposing a mild constraint on QED scores helps guide the generation process toward molecules with a more favorable balance of properties, preventing the reward function from being dominated solely by docking scores or QED values. To normalize the docking score into the range [0,1], we introduce a rescaling factor 
$k$, enabling better compatibility between different reward components. The full definition of the processed docking score is as follows:
\begin{equation}
Dock(s)=\begin{cases}
\frac{\max\left(LeDock(s), k\right)}{k}, &\text{if } s \text{ is \underline{valid}}  \\
0, &\text{otherwise}
\end{cases}
\end{equation}


The agent is trained to learn a policy $\pi(\theta)$ that maximizes the reward expectation, where $\theta$ represents the parameters of the agent model. Our approach builds upon the REINVENT framework~\cite{loeffler2024reinvent}, which has been shown effective in molecular generation tasks, with modifications tailored to our task. To improve training efficiency and stabilize gradient estimation, we adopt a mini-batch training strategy. At each step, a batch of $N$ molecules is sampled from the agent. For a SMILES sequence $s$ generated from $\pi_{\theta}$, we compute its log-likelihood $\log p^{\mathrm{(agent)}}(s)$ under the agent's policy. We also obtain its prior log-likelihood $\log p^{\mathrm{(prior)}}(s)$ by passing the same sequence into the prior model. The objective of RL is to maximize both the prior log-likelihood and reward. Accordingly, we adopt the \underline{p}olicy \underline{g}radient algorithm to maximize the objective and thus the loss function is defined as follows:
\begin{equation}
    \mathcal{L}_{\mathrm{pg}}=-\mathbb{E}_S[\log p^{\mathrm{(agent)}}(s;\theta)\cdot Reward(s)]
\end{equation}

\begin{table*}[!hbpt]
    \centering
\begin{tabular}{|c|c|c|c|c|c|c|c|c|c|c|c|c|}\hline
        & \multicolumn{3}{|c|}{XMolRL }& \multicolumn{3}{|c|}{SmilesGEN}& \multicolumn{3}{|c|}{GxVAEs }& \multicolumn{3}{|c|}{TRIOMPHE}\\\hline
        & Affinity & QED & SA & Affinity & QED & SA & Affinity & QED & SA & Affinity & QED & SA  \\\hline 
        AKT1 & \textbf{-6.49}& \textbf{0.764}& \textbf{2.634} & -5.23 & 0.585 & 2.932 & -5.77 & 0.583 & 3.207 & -2.33 & 0.462 & 4.08  \\\hline 
        AKT2 & \textbf{-6.34} & \textbf{0.729} & \textbf{2.844} & -5.84 & 0.610 & 3.06 & -6.16 & 0.595 & 3.42 & -2.95 & 0.445 & 4.25  \\\hline 
        AURKB & -6.71 &\textbf{ 0.736 }& \textbf{2.612} & -6.21 & 0.574 & 3.01 & \textbf{ -6.94}& 0.599 & 3.276 & -2.31 & 0.425 & 4.10  \\\hline 
        CTSK & -5.07 &\textbf{ 0.675} & 3.00 & -4.81 & 0.604 & \textbf{ 2.97}& \textbf{ -5.33}& 0.597 & 3.011 & -1.87 & 0.399 & 4.47  \\\hline 
        HDAC1 &\textbf{ -4.64} & \textbf{0.729} &\textbf{ 3.197} & -3.51 & 0.566 & 3.017 & -3.59 & 0.619 & 3.225 & -1.84 & 0.437 & 4.30  \\\hline 
        MTOR & -6.27 & \textbf{0.730} & \textbf{2.685} & -5.96 & 0.57 & 3.00 & \textbf{ -6.67}& 0.605 & 3.324 & -2.75 & 0.465 & 4.18  \\\hline 
        PIK3CA & \textbf{-6.04} & \textbf{0.731} & 3.132 & -5.65 & 0.578 & \textbf{ 2.98}& -5.86 & 0.640 & 3.298 & -2.88 & 0.435 & 4.50  \\\hline 
        SMAD3 & \textbf{-5.16 }& \textbf{0.676} & \textbf{3.212 }& -3.54 & 0.577 & 3.01 & -3.4 & 0.594 & 3.242 & -2.57 & 0.464 & 4.15  \\\hline 
        TP53 & \textbf{-7.01} &\textbf{ 0.755 }&\textbf{ 2.81} & -4.56 & 0.593 & 2.97 & -4.38 & 0.611 & 3.281 & -2.64 & 0.417 & 4.17  \\\hline 
        EGFR & \textbf{-7.37} & \textbf{0.726} & \textbf{2.464} & -6.67 & 0.57 & 2.95 & -7.33 & 0.58 & 3.58 & -3.31 & 0.442 & 4.37 \\ \hline 
    \end{tabular}
     \caption{Performance Comparison of XMolRL versus Phenotype-Guided Methods on Affinity, QED and SA Metrics} \label{tab:phenotype}
\end{table*}

\subsection{Ranking Loss and Regularizers}
Reinforcement learning in molecular generation often faces two key challenges: 1) sparse reward signals, which hinder effective policy learning, and 2) unstable sample quality, where the agent may exploit the reward function by generating low-quality yet high-reward molecules. To mitigate these issues and encourage the model to assign higher generation probabilities to molecules with superior properties (e.g., binding affinity, QED, or biological activity), we introduce a ranking loss constraint. Let the desired property score of a molecule be denoted as $AS(s)$, which is defined based on docking scores and relevant chemical properties. Given a pair of molecules $(s_i, s_j)$ with $AS(s_i)>AS(s_j)$, we expect that the agent model assigns higher likelihood probability to the molecule with higher property score. Specifically, the agent's policy $\pi$ is expected to satisfy the ranking constraint
$p^{\mathrm{(agent)}}(s_i|\pi(\theta))>p^{\mathrm{(agent)}}(s_j|\pi(\theta))$ for all $AS(s_i)>AS(s_j)$.
When a lower-quality molecule is assigned a higher generative probability than a higher-quality one, a ranking loss is incurred as below:
\begin{align}
\mathcal{L}_{\mathrm{rank}}=\sum_i\sum_{j>i}\max\left(0,f(s_j)-f(s_i)+\gamma_{ij}\right), \nonumber \\ 
\forall i<j, AS(s_i)>AS(s_j)
\end{align}
where $\gamma_{ij}=(j-i)\cdot\gamma$ is defined as the rank difference between two molecules multiplied by a hyperparameter $\gamma$. The function $f(s)$ is defined as $\log p^{\mathrm{(agent)}}(s;\theta)$. 
By increasing the generative probability of high-quality molecules, RL no longer relies solely on sparse rewards, but instead incorporates fine-grained feedback through property-based ranking. This helps prevent the agent from overfitting to syntactically valid but chemically poor molecules. Compared to purely reward-driven optimization, the use of rank loss enables the model to effectively distinguish structurally similar molecules with significantly different properties, thereby enhancing its structural optimization capability.

To ensure the stability and effectiveness of RL, we introduce two additional regularizers: prior likelihood regularization~\cite{gomez2018automatic,loeffler2024reinvent} and entropy regularization~\cite{you2018graph,zoph2016neural}. To prevent the agent from deviating from the prior model, we incorporate a prior likelihood term based on the phenotype-guided generator. This regularizer is defined as:
\begin{equation}
\mathcal{L}_{\text{prior}} = -\mathbb{E}_S[\log p^{\text{(prior)}}(s)]
\end{equation}

Here, $\log p_{\text{prior}}(s)$ denotes the log-likelihood of a molecule $s$ under the prior model. The prior poses the constraint that the generated molecules conform to valid SMILES syntax and chemically plausible structures, while also satisfying phenotype conditions with high fidelity. By imposing this prior constraint, the agent model benefits from the inductive biases encoded in the trained phenotypic generator, which improves its ability to respond accurately to phenotypic signatures. This is especially crucial in the early stages of RL, where the prior prevents the agent from cheating the reward  function (``reward hackin'') by generating unrealistic or syntactically invalid molecules.

To encourage chemical space exploration and increase structural diversity, we introduce an entropy regularization term, which is defined as:
\begin{equation}
\mathcal{L}_{\text{ent}} = \mathbb{E}_S \left[ H(s;\theta) \right]
\end{equation}
in which $ H(s;\theta)=-\sum_{t=1}^T  p(a | a_{<t}; \theta)\cdot\log p(a | a_{<t}; \theta)$ denotes the entropy of the agent's action distribution over the next token. Maximizing this entropy encourages stochasticity in the policy, helping the agent avoid deterministic local optima and promoting molecular diversity.

Taking the ranking loss and two regularization terms into the policy gradient loss, the overall objective enables the joint optimization of  molecular quality while preserving generative capacity. The final loss function is defined as:
\begin{equation}
\mathcal{L} = \mathcal{L}_{\text{pg}} + \alpha \cdot \mathcal{L}_{\text{rank}} + \beta \cdot \mathcal{L}_{\text{prior}} - \lambda \cdot \mathcal{L}_{\text{ent}} 
\end{equation}
where $\alpha$, $\beta$, and $\lambda$ are hyperparameters used to balance the weights of the ranking loss, prior and entropy regularization, respectively. In this study, they are empirically tuned to 0.2, 0.8, and 0.1, respectively.

In our practice, we employed the bidirectional GRUs as the backbone of MolVAE. Both the encoder and decoder were composed of three hidden layers of size 192. The ExpVAE adopted multiple feed-forward layers with sizes 512, 256, and 192, respectively. During the training of phenotypic generator, the Adam optimizer is used with a learning rate of 5e-4 and a dropout rate of 0.1. The maximum length of the generated SMILES strings was restricted to 100 characters. In the RL stage, agent model was optimized using Adam optimizer with a reduced learning rate of 1e-4. The batch size for both stages was set at 64. Our model was implemented using PyTorch 2.3.0. The training and all evaluation experiments were conducted on a CentOS Linux 8.2.2004 (Core) system, equipped with a GeForce RTX 4090 GPU and 128GB of memory.  



\begin{figure*}[htb]
        \centering
        \includegraphics[width=1\linewidth]{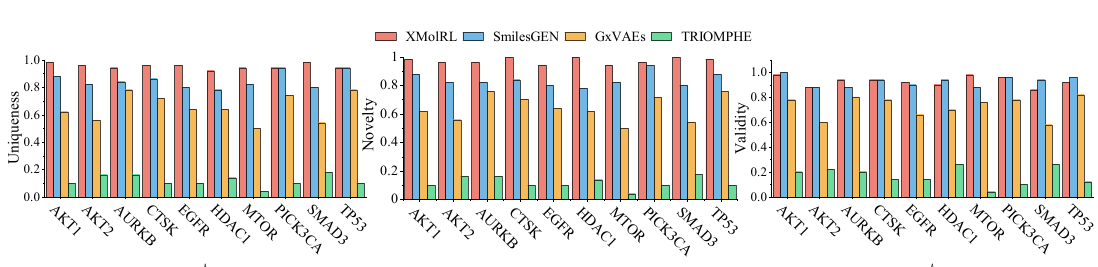}
        \caption{Performance Comparison of XMolRL to Phenotype-Guided Methods on Uniqueness, Novelty and Validity}
        \label{fig:phenotype}
\end{figure*}

\section{Experiments}
\subsection{Datasets and Preprocessing}
The MolVAE was pretrained using 10,032,879 SMILES sequences obtained from the ZINC database~\cite{irwin2005zinc}, and then the ExpVAE and MolVAE were jointly trained using
the L1000 dataset \cite{SUBRAMANIAN20171437}, which was a collective repository of transcriptional responses of human cell lines to drug exposures. We extracted the expression profiles of cell lines treated by 10$\mu$M drug concentration for 24 hours. To streamline the data, technical replicates were averaged. As a result, we obtained drug-induced expression profiles spanning 6,549 drugs and 164 cell lines. The refined dataset comprised a total of 86,400 expression profiles across 978 landmark genes to fine-tune the dual VAEs. In addition, to evaluate the binding affinity of generated molecules toward target proteins, we collected co-crystal structures of the ten protein–ligand complexes from the RCSB Protein Data Bank. Protein structures were preprocessed using the LePro tool (http://www.lephar.com/), and molecular docking was performed with the LeDock tool. 
For performance evaluation, we used the expression profiles induced by the genetic perturbation of ten commonly target genes associated with the treatment of cancers. Among them, eight genes (AKT1, AKT2, AURKB, CTSK, EGFR, HDAC1, MTOR, and PIK3CA) were knockdown, and two genes (SMAD3 and TP53) were over-expressed. To evaluate the generated molecules, we obtained known ligands that have been confirmed to target these proteins from the Drug Target Commons (DTC) database~\cite{tang2018drug}. We computed the performance metrics between these known ligands and generated molecules.

\subsection{Performance Metrics}
To evaluate the chemical properties of the generated molecules, we measured quite a few key metrics: validity, uniqueness, novelty, quantitative estimate of drug-likeness (QED), and synthetic accessibility (SA).  High validity and uniqueness are indicators of an effective molecule generation process, while high novelty indicates the model avoids overfitting to the training set. For assessing binding affinity to target protein, we employed the docking scores computed by LeDock. In particular, we adopted IC50 as a key performance metric for evaluating phenotype-guided generative models. IC50 represents the concentration of a compound required to inhibit 50\% of cancer cell viability, with lower values indicating greater inhibitory potency. In our experiments, IC50 values were predicted using the PaccMann platform (https://ibm.biz/paccmann-aas)~\cite{cadow2020paccmann}. These performance metrics ensure that our model was comprehensively evaluated in generating molecules with potential therapeutic value and practical feasibility.


\begin{table*}[!ht]
    \centering
    \begin{tabular}{|c|c|c|c|c|c|c|c|c|c|c|c|c|}\hline
    
        & \multicolumn{3}{|c|}{XMolRL}& \multicolumn{3}{|c|}{Pocket2Mol }& \multicolumn{3}{|c|}{SBDD-3D}&  \multicolumn{3}{|c|}{SBMolGen }\\\hline
        & Affinity & QED & SA & Affinity & QED & SA & Affinity & QED & SA & Affinity & QED & SA  \\\hline 
        AKT1 & \textbf{-6.49}& \textbf{0.764} & \textbf{2.634}& -2.82 & 0.436 & 2.622 & -2.82& 0.525& 3.26& -4.76 & 0.722 & 2.897  \\\hline 
        AKT2 & -6.34& \textbf{0.729}& \textbf{2.844}& -5.99 & 0.612 & 3.362 & \textbf{-6.99}& 0.399& 5.42& -5.49 & 0.710 & 3.042  \\\hline 
        AURKB & -6.71& 0.736 & 2.612 & -5.47 & 0.687 & \textbf{2.600}& \textbf{-6.83}& 0.566& 5.36& -5.91 & \textbf{ 0.737}& 2.927  \\\hline 
        CTSK & \textbf{-5.07} & 0.675 & \textbf{3.00}& -4.67 & \textbf{0.723}& 2.803 & -4.55 & 0.672& 4.98& -4.61 & 0.713 & 3.032  \\\hline 
        HDAC1 & \textbf{-4.64} & \textbf{0.729} & 3.197 & -2.35 & 0.507 & \textbf{2.535}& -2.83& 0.618& 2.86& -3.54 & 0.713 & 2.95  \\\hline 
        MTOR & \textbf{-6.27} & \textbf{0.730}& \textbf{2.685}& -5.68 & 0.639 & 3.189 & -5.22& 0.559&5.748& -5.43 & 0.727 & 2.970  \\\hline 
        PIK3CA & \textbf{-6.04} & 0.731& 3.132 & -4.70 & \textbf{ 0.750 }& \textbf{2.826}& -4.47& 0.612& 4.25& -5.53 & 0.706 & 2.936  \\\hline 
        SMAD3 & \textbf{-5.16} & 0.676& 3.212 & -3.72 & 0.598 & \textbf{ 2.459} & -3.31& 0.385& 3.91& -3.92 & \textbf{0.747}& 3.091  \\\hline 
        TP53 & \textbf{-7.01} & \textbf{0.755} & \textbf{2.81}& -5.25 & 0.656 & 3.351 & -5.16& 0.573& 5.127& -5.51 & 0.709 & 2.992  \\\hline 
        EGFR & \textbf{-7.37} & 0.726 & \textbf{2.464}& -7.02 & 0.610 & 3.469 & -6.32& 0.566& 5.68& -6.14 & \textbf{ 0.737} & 3.062 \\ \hline 
    \end{tabular}
        \caption{Performance Comparison of XMolRL versus Target-Based Methods on Affinity, QED and SA Metrics}\label{tab:target}
\end{table*}

\subsection{Comparison to Phenotype-Guided Methods}
We first compare XMolRL with several representative phenotype-based molecular generation methods, including SmilesGEN~\cite{liu2025phenotypic}, GxVAEs~\cite{li2024gxvaes}, and TRIOMPHE~\cite{kaitoh2021triomphe}. For each target protein, we generated 100 unique molecules using these methods and evaluated their performance across multiple metrics, as shown in Table~\ref{tab:phenotype}. First, XMolRL achieved the best docking scores on 7 out of the 10 target proteins, with the generated molecules exhibiting significantly stronger binding affinities compared to those produced by SmilesGEN, GxVAEs, and TRIOMPHE. This highlights our model's superior capability in optimizing chemical structure for specific binding pockets. Although GxVAEs performed comparably or slightly better on a few targets (e.g., AURKB and MTOR), its overall performance was inferior to our method, suggesting that our method offers better generalization and robustness across diverse protein contexts. 

Moreover, XMolRL consistently achieved the highest QED scores across all ten target proteins, indicating that the generated molecules exhibit more drug-like pharmacological properties. Notably, even though SA scores were not explicitly included in the RL reward function, XMolRL still outperformed the competing methods on most targets with respect to the SA metric. We further analyzed the distribution of molecular property metrics across different methods. As shown in Figure S1, XMolRL achieved substantially higher affinity than TRIOMPHE and SmilesGEN, with peaks concentrated in the high-affinity region (lower score means higher affinity). Meanwhile, XMolRL consistently outperformed all competing methods in QED and SA score distributions, demonstrating its superior ability to generate drug-like and synthetically accessible compounds. To further validate the structural fidelity, we calculated the maximum Tanimoto similarity between the generated molecules and known ligands for each target. As illustrated in Figure S2, the generated molecules achieving the highest similarity scores are displayed alongside their corresponding reference ligands, verifying XMolRL's capability to accurately reproduce key structural features of active compounds. 

Furthermore, we evaluated the uniqueness, novelty and validity of our generated molecules. As shown in Figure~\ref{fig:phenotype}, XMolRL outperformed four competing methods across all 10 targets, with novelty reaching 100\% on targets such as TP53. These results demonstrate that XMolRL consistently generates structurally novel compounds unseen during training. 

\begin{figure}[hptb]
    \centering
    \includegraphics[width=1\linewidth]{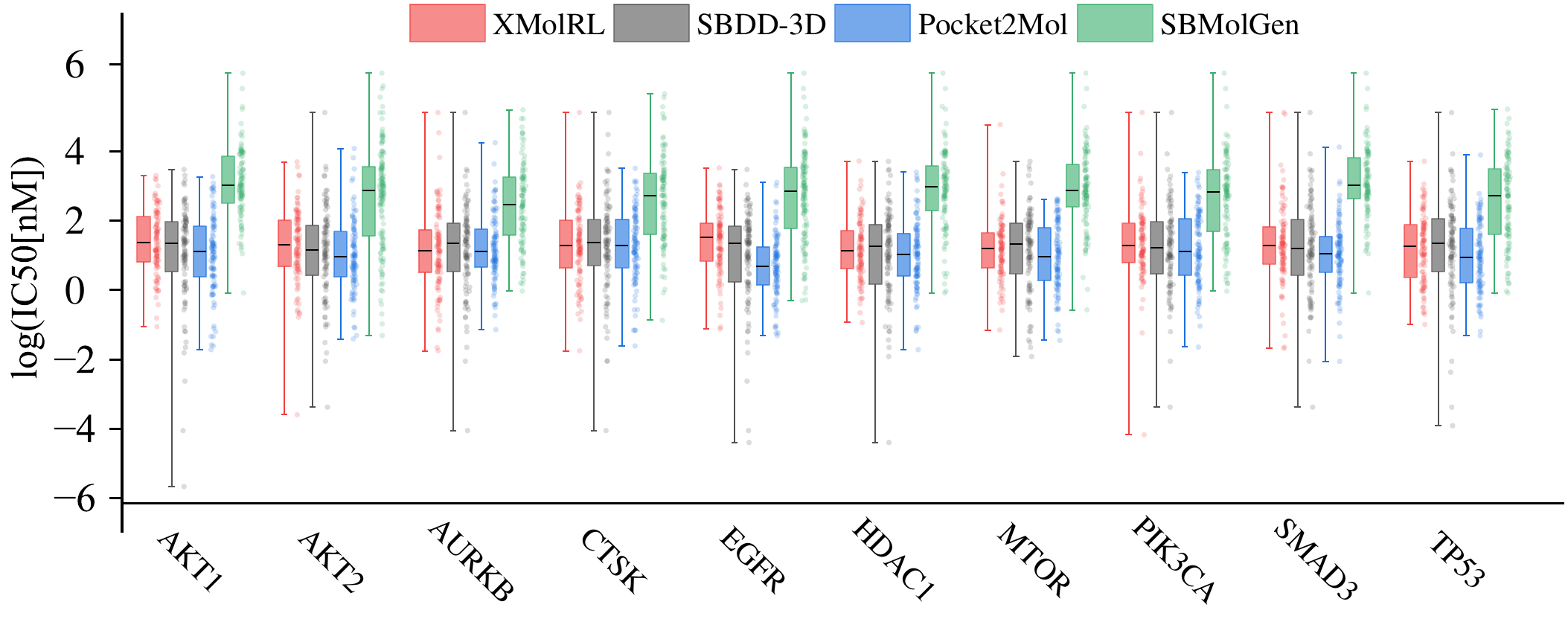}
    \caption{Performance comparison of XMolRL versus targeted approaches in suppressing cancer cell viability}
    \label{fig:IC50}
\end{figure}

\begin{figure*}[htb]
        \centering
        \includegraphics[width=1\linewidth]{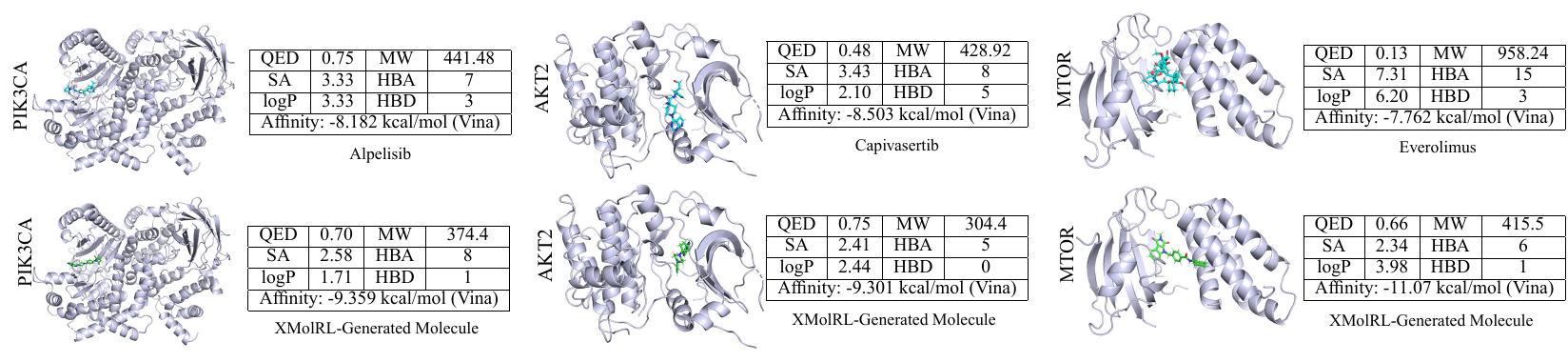}
    \caption{Comparison of XMolRL‑generated molecules versus approved drugs for the PIK3CA, AKT2, and mTOR targets. }
        \label{fig:approved}
\end{figure*}

\subsection{Comparison to Target-Based Methods}
We next compared XMolRL with three docking-based generative methods: SBDD-3D~\cite{luo20213d},Pocket2Mol~\cite{peng2022pocket2mol}, and SBMolGen~\cite{ma2021structure}. Similarly, we generated 100 unique molecules using each method for performance evaluation. As shown in Table~\ref{tab:target}, XMolRL outperformed the competing methods in both docking scores and QED. 
We used EGFR as a representative target to analyze the distribution of evaluation metrics. As shown in Figure S3, XMolRL exhibited substantial advantage over competing methods in both docking and QED scores, with peaks concentrated in the advantageous region. These results demonstrate that XMolRL effectively leveraged both phenotypic guidance and target-specific docking signals, leading to superior molecular generation performance across multiple evaluation metrics. Furthermore, we visualized the docking poses of the top-performing candidate molecules for each target (Figure S4), which indicates that the generated ligands adopt stable and reasonable conformations within the active sites and form effective interactions with key residues.  

Target-based methods focus solely on generating high-affinity molecules to specific target proteins, overlooking the effect on cellular contexts. In contrast, XMolRL is explicitly conditioned on the cellular state of the target protein to generate molecules that would induce desirable phenotypic profiles. To assess the effectiveness of this strategy, we further evaluated the ability of generated molecules to inhibit the viability of MCF7 breast cancer cells. Figure~\ref{fig:IC50} showed the boxplots of the log(IC50) values that measure the inhibitory potency of molecules generated by XMolRL and competing methods across 10 target proteins. Note that lower log(IC50) values indicate stronger inhibitory capacity. The results verified that XMolRL consistently produced molecules with stronger anticancer potential across all targets compared to other methods. Moreover, the IC50 distributions of XMolRL exhibited lower variance, indicating its stability to generate high-quality molecules. While SBMolGen did generate a few highly potent molecules with extremely low log(IC50) values for specific targets such as AKT2 and SMAD, its overall distribution was broader and less stable. These results highlight that, compared to docking-based approaches, XMolRL's phenotype-conditioned strategy enables more effective and personalized molecule design, particularly for targeting cancer-specific cellular states.

\subsection{Therapeutic Molecular Generation}
To assess XMolRL's ability to generate therapeutic compounds, we produced 100 novel molecules conditioned on breast cancer expression profiles and targeting PIK3CA, AKT2, and MTOR. These molecules were benchmarked against three FDA-approved drugs: Alpelisib (PI3K$\alpha$ inhibitor), Capivasertib (AKT inhibitor), and Everolimus (mTOR inhibitor). All candidates were firstly filtered according to Lipinski's Rule (LogP$<$5, MW$<$500Da, H‑bond donors$<$5, H‑bond acceptors$<$10), and then were docked to the target proteins using AutoDock Vina~\cite{trott2010autodock}. Other key physicochemical properties, including QED and synthetic accessibility (SA), were also computed for further evaluation. As shown in Figure~\ref{fig:approved}, we illustrate the docking conformation of the three approved drugs (upper) and XMolRL-generated candidates (bottom), alongside their QED, SA, and Vina scores. XMolRL‑generated molecules consistently outperformed the known drugs in binding affinity, achieving Vina scores of -9.359, -9.301, and -11.070 kcal/mol versus -8.182, -8.503, and -7.762 kcal/mol for Alpelisib, Capivasertib, and Everolimus, respectively. These candidates also exhibited improved drug‑likeness and SA scores. Taking mTOR as example, Everolimus scored a QED of only 0.13 with an SA score of 7.31, whereas XMolRL-generated candidate achieved a QED of 0.66 and an SA score of 2.34. Moreover, the generated candidate's molecular weight (415.5 Da) and Log P align more closely with Lipinski criteria than Everolimus (958.2 Da), suggesting enhanced absorption and bioavailability. These results demonstrate that XMolRL not only generates novel compounds with superior target binding, but also optimizes critical physicochemical properties.

\subsection{Ablation Study}
To evaluate the contribution of key components in XMolRL, we conducted ablation studies on the EGFR target by removing the ranking loss or replacing multi-objective optimization with single-objective optimization (Affinity or QED) (Table~\ref{ablation}). The results show that the ranking loss is crucial for maintaining molecular uniqueness: its removal reduces uniqueness from 96\% to 56\%, indicating severe mode collapse and metric inflation due to repeated sampling. Single-objective optimization also exhibits clear bias—optimizing affinity alone degrades QED and synthetic accessibility (SA), while optimizing QED alone substantially compromises binding affinity. In contrast, the full XMolRL model effectively balances diversity, drug-likeness, and synthetic feasibility without sacrificing docking performance, resulting in more practically valuable candidate molecules.

\begin{table}[!ht]
\centering
\label{tab:ablation}
\setlength{\tabcolsep}{1pt} 
\renewcommand{\arraystretch}{1.2} 
\begin{tabular}{|l|c|c|c|c|}
\hline
\textbf{Metric} 
& \textbf{XMolRL}& \textbf{w/o Rank}& \textbf{w/o QED}& \textbf{w/o Affinity}\\
\hline
Uniqueness (\%)        & 96.0& 56.0& 76.0& 98.0\\\hline
Validity (\%)          & 100.0& 98.0& 96.0& 90.0\\\hline
Novelty (\%)           & 94.0& 96.0& 100.0& 98.0\\\hline
Affinity ($\downarrow$)& -7.53 & -6.89 & -7.54 & -5.35 \\\hline
QED ($\uparrow$)       & 0.717 & 0.748 & 0.504 & 0.770 \\\hline
SA ($\downarrow$)      & 2.54  & 2.63  & 3.01  & 2.72 \\
\hline
\end{tabular}
\caption{Performance achieved by different ablated models} \label{ablation}
\end{table}

\section{Conclusion}

The proposed XMolRL framework integrates phenotypic profiles and target protein structures via multi-objective reinforcement learning to generate de novo molecules with enhanced efficacy and specificity. Extensive experiments demonstrate its superiority over existing methods in drug-likeness, binding affinity, and phenotypic relevance, highlighting its potential for drug design.

\bibliographystyle{named}
\bibliography{ijcai26}

@article{das2023gex2sgen,
  title={Gex2SGen: designing drug-like molecules from desired gene expression signatures},
  author={Das, Dibyajyoti and Chakrabarty, Broto and Srinivasan, Rajgopal and Roy, Arijit},
  journal={Journal of Chemical Information and Modeling},
  volume={63},
  number={7},
  pages={1882--1893},
  year={2023},
  publisher={ACS Publications}
}

@inproceedings{li2024gxvaes,
  title={GxVAEs: Two Joint VAEs Generate Hit Molecules from Gene Expression Profiles},
  author={Li, Chen and Yamanishi, Yoshihiro},
  booktitle={Proceedings of the AAAI Conference on Artificial Intelligence},
  pages={13455--13463},
  year={2024}
}

@article{kaitoh2021triomphe,
  title={TRIOMPHE: transcriptome-based inference and generation of molecules with desired phenotypes by machine learning},
  author={Kaitoh, Kazuma and Yamanishi, Yoshihiro},
  journal={Journal of Chemical Information and Modeling},
  volume={61},
  number={9},
  pages={4303--4320},
  year={2021},
  publisher={ACS Publications}
}

@article{vincent2022phenotypic,
  title={Phenotypic drug discovery: recent successes, lessons learned and new directions},
  author={Vincent, Fabien and Nueda, Arsenio and Lee, Jonathan and Schenone, Monica and Prunotto, Marco and Mercola, Mark},
  journal={Nature Reviews Drug Discovery},
  volume={21},
  number={12},
  pages={899--914},
  year={2022},
  publisher={Nature Publishing Group UK London}
}

@article{meissner2022emerging,
  title={The emerging role of mass spectrometry-based proteomics in drug discovery},
  author={Meissner, Felix and Geddes-McAlister, Jennifer and Mann, Matthias and Bantscheff, Marcus},
  journal={Nature Reviews Drug Discovery},
  volume={21},
  number={9},
  pages={637--654},
  year={2022},
  publisher={Nature Publishing Group UK London}
}

@article{gomez2018automatic,
  title={Automatic chemical design using a data-driven continuous representation of molecules},
  author={G{\'o}mez-Bombarelli, Rafael and Wei, Jennifer N and Duvenaud, David and Hern{\'a}ndez-Lobato, Jos{\'e} Miguel and S{\'a}nchez-Lengeling, Benjam{\'\i}n and Sheberla, Dennis and Aguilera-Iparraguirre, Jorge and Hirzel, Timothy D and Adams, Ryan P and Aspuru-Guzik, Al{\'a}n},
  journal={ACS central science},
  volume={4},
  number={2},
  pages={268--276},
  year={2018},
  publisher={ACS Publications}
}

@article{SUBRAMANIAN20171437,
title = {A Next Generation Connectivity Map: L1000 Platform and the First 1,000,000 Profiles},
journal = {Cell},
volume = {171},
number = {6},
pages = {1437-1452.e17},
year = {2017},
author = {Aravind Subramanian and Rajiv Narayan and Steven M. Corsello and et al.}
}

@article{tang2018drug,
  title={Drug target commons: a community effort to build a consensus knowledge base for drug-target interactions},
  author={Tang, Jing and Ravikumar, Balaguru and Alam, Zaid and Rebane, Anni and V{\"a}h{\"a}-Koskela, Markus and Peddinti, Gopal and van Adrichem, Arjan J and Wakkinen, Janica and Jaiswal, Alok and Karjalainen, Ella and others},
  journal={Cell chemical biology},
  volume={25},
  number={2},
  pages={224--229},
  year={2018},
  publisher={Elsevier}
}

@article{Subramanian2017L1000,
  title={A Next Generation Connectivity Map: L1000 Platform and the First 1,000,000 Profiles},
  author={Subramanian, Aravind and Narayan, Rajiv and Corsello, Samantha M and others},
  journal={Cell},
  volume={171},
  number={6},
  pages={1437--1452.e17},
  year={2017},
  publisher={Elsevier}
}

@inproceedings{peng2022pocket2mol,
  title={Pocket2Mol: Efficient Molecular Sampling Based on 3D Protein Pockets},
  author={Xingang Peng and Shitong Luo and Jiaqi Guan and Qi Xie and Jian Peng and Jianzhu Ma},
  booktitle={International Conference on Machine Learning},
  year={2022}
}

@article{xu2021navigating,
  title={Navigating chemical space by interfacing generative artificial intelligence and molecular docking},
  author={Xu, Ziqiao and Wauchope, Orrette R and Frank, Aaron T},
  journal={Journal of Chemical Information and Modeling},
  volume={61},
  number={11},
  pages={5589--5600},
  year={2021},
  publisher={ACS Publications}
}

@article{ma2021structure,
  title={Structure-based de novo molecular generator combined with artificial intelligence and docking simulations},
  author={Ma, Biao and Terayama, Kei and Matsumoto, Shigeyuki and Isaka, Yuta and Sasakura, Yoko and Iwata, Hiroaki and Araki, Mitsugu and Okuno, Yasushi},
  journal={Journal of Chemical Information and Modeling},
  volume={61},
  number={7},
  pages={3304--3313},
  year={2021},
  publisher={ACS Publications}
}

@article{loeffler2024reinvent,
  title={Reinvent 4: modern AI--driven generative molecule design},
  author={Loeffler, Hannes H and He, Jiazhen and Tibo, Alessandro and Janet, Jon Paul and Voronov, Alexey and Mervin, Lewis H and Engkvist, Ola},
  journal={Journal of Cheminformatics},
  volume={16},
  number={1},
  pages={20},
  year={2024},
  publisher={Springer}
}

@article{cadow2020paccmann,
  title={PaccMann: a web service for interpretable anticancer compound sensitivity prediction},
  author={Cadow, Joris and Born, Jannis and Manica, Matteo and Oskooei, Ali and Rodr{\'\i}guez Mart{\'\i}nez, Mar{\'\i}a},
  journal={Nucleic acids research},
  volume={48},
  number={W1},
  pages={W502--W508},
  year={2020},
  publisher={Oxford University Press}
}

@article{aini2024antimalarial,
  title={Antimalarial Potential of Phytochemical Compounds from Garcinia atroviridis Griff ex. T. Anders Targeting Multiple Proteins of Plasmodium falciparum 3D7: An In Silico Approach},
  author={Aini, Nur Sofiatul and Ansori, Arif Nur Muhammad and Herdiansyah, Mochammad Aqilah and Kharisma, Viol Dhea and Widyananda, Muhammad Hermawan and Murtadlo, Ahmad Affan Ali and Turista, Dora Dayu Rahma and Sucipto, Teguh Hari and Sahadewa, Sukma and Durry, Fara Disa and others},
  journal={BIO Integration},
  volume={5},
  number={1},
  pages={967},
  year={2024},
  publisher={Compuscript}
}

@article{danel2023docking,
  title={Docking-based generative approaches in the search for new drug candidates},
  author={Danel, Tomasz and Leski, Jan and Podlewska, Sabina and Podolak, Igor T},
  journal={Drug Discovery Today},
  volume={28},
  number={2},
  pages={103439},
  year={2023},
  publisher={Elsevier}
}

@article{pang2023deep,
  title={Deep generative models in de novo drug molecule generation},
  author={Pang, Chao and Qiao, Jianbo and Zeng, Xiangxiang and Zou, Quan and Wei, Leyi},
  journal={Journal of Chemical Information and Modeling},
  volume={64},
  number={7},
  pages={2174--2194},
  year={2023},
  publisher={ACS Publications}
}

@article{sanchez2018inverse,
  title={Inverse molecular design using machine learning: Generative models for matter engineering},
  author={Sanchez-Lengeling, Benjamin and Aspuru-Guzik, Al{\'a}n},
  journal={Science},
  volume={361},
  number={6400},
  pages={360--365},
  year={2018},
  publisher={American Association for the Advancement of Science}
}

@article{brown2019guacamol,
  title={GuacaMol: benchmarking models for de novo molecular design},
  author={Brown, Nathan and Fiscato, Marco and Segler, Marwin HS and Vaucher, Alain C},
  journal={Journal of chemical information and modeling},
  volume={59},
  number={3},
  pages={1096--1108},
  year={2019},
  publisher={ACS Publications}
}

@article{hughes2011principles,
  title={Principles of early drug discovery},
  author={Hughes, James P and Rees, Stephen and Kalindjian, S Barrett and Philpott, Karen L},
  journal={British journal of pharmacology},
  volume={162},
  number={6},
  pages={1239--1249},
  year={2011},
  publisher={Wiley Online Library}
}

@article{imming2006drugs,
  title={Drugs, their targets and the nature and number of drug targets},
  author={Imming, Peter and Sinning, Christian and Meyer, Achim},
  journal={Nature reviews Drug discovery},
  volume={5},
  number={10},
  pages={821--834},
  year={2006},
  publisher={Nature Publishing Group UK London}
}

@article{wishart2018drugbank,
  title={DrugBank 5.0: a major update to the DrugBank database for 2018},
  author={Wishart, David S and Feunang, Yannick D and Guo, An C and Lo, Elvis J and Marcu, Ana and Grant, Jason R and Sajed, Tanvir and Johnson, Daniel and Li, Carin and Sayeeda, Zinat and others},
  journal={Nucleic acids research},
  volume={46},
  number={D1},
  pages={D1074--D1082},
  year={2018},
  publisher={Oxford University Press}
}

@article{swinney2011were,
  title={How were new medicines discovered?},
  author={Swinney, David C and Anthony, Jason},
  journal={Nature reviews Drug discovery},
  volume={10},
  number={7},
  pages={507--519},
  year={2011},
  publisher={Nature Publishing Group UK London}
}

@article{polykovskiy2020molecular,
  title={Molecular sets (MOSES): a benchmarking platform for molecular generation models},
  author={Polykovskiy, Daniil and Zhebrak, Alexander and Sanchez-Lengeling, Benjamin and Golovanov, Sergey and Tatanov, Oktai and Belyaev, Stanislav and Kurbanov, Rauf and Artamonov, Aleksey and Aladinskiy, Vladimir and Veselov, Mark and others},
  journal={Frontiers in pharmacology},
  volume={11},
  pages={565644},
  year={2020},
  publisher={Frontiers Media SA}
}

@article{von2020exploring,
  title={Exploring chemical compound space with quantum-based machine learning},
  author={von Lilienfeld, O Anatole and M{\"u}ller, Klaus-Robert and Tkatchenko, Alexandre},
  journal={Nature Reviews Chemistry},
  volume={4},
  number={7},
  pages={347--358},
  year={2020},
  publisher={Nature Publishing Group UK London}
}

@article{moffat2014phenotypic,
  title={Phenotypic screening in cancer drug discovery—past, present and future},
  author={Moffat, John G and Rudolph, Joachim and Bailey, David},
  journal={Nature reviews Drug discovery},
  volume={13},
  number={8},
  pages={588--602},
  year={2014},
  publisher={Nature Publishing Group UK London}
}

@article{moffat2017opportunities,
  title={Opportunities and challenges in phenotypic drug discovery: an industry perspective},
  author={Moffat, John G and Vincent, Fabien and Lee, Jonathan A and Eder, J{\"o}rg and Prunotto, Marco},
  journal={Nature reviews Drug discovery},
  volume={16},
  number={8},
  pages={531--543},
  year={2017},
  publisher={Nature Publishing Group UK London}
}

@article{lamb2006connectivity,
  title={The Connectivity Map: using gene-expression signatures to connect small molecules, genes, and disease},
  author={Lamb, Justin and Crawford, Emily D and Peck, David and Modell, Joshua W and Blat, Irene C and Wrobel, Matthew J and Lerner, Jim and Brunet, Jean-Philippe and Subramanian, Aravind and Ross, Kenneth N and others},
  journal={science},
  volume={313},
  number={5795},
  pages={1929--1935},
  year={2006},
  publisher={American Association for the Advancement of Science}
}

@article{you2018graph,
  title={Graph convolutional policy network for goal-directed molecular graph generation},
  author={You, Jiaxuan and Liu, Bowen and Ying, Zhitao and Pande, Vijay and Leskovec, Jure},
  journal={Advances in neural information processing systems},
  volume={31},
  year={2018}
}

@article{zoph2016neural,
  title={Neural architecture search with reinforcement learning},
  author={Zoph, Barret and Le, Quoc V},
  journal={arXiv preprint arXiv:1611.01578},
  year={2016}
}

@article{mendez2020novo,
  title={De novo generation of hit-like molecules from gene expression signatures using artificial intelligence},
  author={M{\'e}ndez-Lucio, Oscar and Baillif, Benoit and Clevert, Djork-Arn{\'e} and Rouqui{\'e}, David and Wichard, Joerg},
  journal={Nature communications},
  volume={11},
  number={1},
  pages={10},
  year={2020},
  publisher={Nature Publishing Group UK London}
}

@article{spiegel2020autogrow4,
  title={AutoGrow4: an open-source genetic algorithm for de novo drug design and lead optimization},
  author={Spiegel, Jacob O and Durrant, Jacob D},
  journal={Journal of cheminformatics},
  volume={12},
  number={1},
  pages={25},
  year={2020},
  publisher={Springer}
}

@article{nigam2021beyond,
  title={Beyond generative models: superfast traversal, optimization, novelty, exploration and discovery (STONED) algorithm for molecules using SELFIES},
  author={Nigam, AkshatKumar and Pollice, Robert and Krenn, Mario and dos Passos Gomes, Gabriel and Aspuru-Guzik, Alan},
  journal={Chemical science},
  volume={12},
  number={20},
  pages={7079--7090},
  year={2021},
  publisher={Royal Society of Chemistry}
}

@article{liu2025phenotypic,
  title={Phenotypic Profile-Informed Generation of Drug-Like Molecules via Dual-Channel Variational Autoencoders},
  author={Liu, Hui and Tian, Shiye and Liu, Xuejun},
  journal={arXiv preprint arXiv:2506.02051},
  year={2025}
}

@article{irwin2005zinc,
  title={ZINC- a free database of commercially available compounds for virtual screening},
  author={Irwin, John J and Shoichet, Brian K},
  journal={Journal of chemical information and modeling},
  volume={45},
  number={1},
  pages={177--182},
  year={2005},
  publisher={ACS Publications}
}

@article{trott2010autodock,
  title={AutoDock Vina: improving the speed and accuracy of docking with a new scoring function, efficient optimization, and multithreading},
  author={Trott, Oleg and Olson, Arthur J},
  journal={Journal of computational chemistry},
  volume={31},
  number={2},
  pages={455--461},
  year={2010},
  publisher={Wiley Online Library}
}

@article{zhao2013discovery,
  title={Discovery of ZAP70 inhibitors by high-throughput docking into a conformation of its kinase domain generated by molecular dynamics},
  author={Zhao, Hongtao and Caflisch, Amedeo},
  journal={Bioorganic \& medicinal chemistry letters},
  volume={23},
  number={20},
  pages={5721--5726},
  year={2013},
  publisher={Elsevier}
}

@article{wang2016comprehensive,
  title={Comprehensive evaluation of ten docking programs on a diverse set of protein--ligand complexes: the prediction accuracy of sampling power and scoring power},
  author={Wang, Zhe and Sun, Huiyong and Yao, Xiaojun and Li, Dan and Xu, Lei and Li, Youyong and Tian, Sheng and Hou, Tingjun},
  journal={Physical Chemistry Chemical Physics},
  volume={18},
  number={18},
  pages={12964--12975},
  year={2016},
  publisher={Royal Society of Chemistry}
}

@article{luo20213d,
  title={A 3D generative model for structure-based drug design},
  author={Luo, Shitong and Guan, Jiaqi and Ma, Jianzhu and Peng, Jian},
  journal={Advances in Neural Information Processing Systems},
  volume={34},
  pages={6229--6239},
  year={2021}
}

@article{bickerton2012quantifying,
  title={Quantifying the chemical beauty of drugs},
  author={Bickerton, G Richard and Paolini, Gaia V and Besnard, J{\'e}r{\'e}my and Muresan, Sorel and Hopkins, Andrew L},
  journal={Nature chemistry},
  volume={4},
  number={2},
  pages={90--98},
  year={2012},
  publisher={Nature Publishing Group UK London}
}

\end{document}